\pdfoutput=1  
\documentclass{vgtc}                

\graphicspath{{figures/}}
\usepackage{times}

\usepackage{mathptmx}
\usepackage{booktabs}
\usepackage{verbatim}
\usepackage{enumitem}

\makeatletter
\g@addto@macro\normalsize{%
  \abovedisplayskip=5pt plus 2pt minus 3pt
  \belowdisplayskip=5pt plus 2pt minus 3pt
  \abovedisplayshortskip=2pt plus 2pt
  \belowdisplayshortskip=3pt plus 2pt}
\AtBeginDocument{%
  \renewcommand\section{\@startsection{section}{1}{\z@}%
                 {-1.6ex \@plus -0.8ex \@minus -.2ex}%
                 {0.6ex \@plus .2ex}%
                 {\reset@font\normalsize\sffamily\bfseries\scshape\vgtc@sectionfont}}%
  \renewcommand\subsection{\@startsection{subsection}{2}{\z@}%
                 {-1.0ex\@plus -0.5ex \@minus -.2ex}%
                 {0.3ex \@plus .2ex}%
                 {\reset@font\normalsize\sffamily\bfseries\vgtc@sectionfont}}%
  \renewcommand\subsubsection{\@startsection{subsubsection}{3}{\z@}%
                 {-1.0ex\@plus -0.5ex \@minus -.2ex}%
                 {0.3ex \@plus .2ex}%
                 {\reset@font\sffamily\normalsize\vgtc@sectionfont}}%
}
\makeatother

\onlineid{0}
\vgtccategory{Research}
\vgtcinsertpkg

\usepackage{amssymb}
\definecolor{famsup}{HTML}{0072B2}   
\definecolor{famssl}{HTML}{E69F00}   
\definecolor{famcvl}{HTML}{009E73}   
\definecolor{famopen}{HTML}{CC79A7}  
\definecolor{famapi}{HTML}{D55E00}   
\newcommand{\ssup}{\textcolor{famsup}{\raisebox{.22ex}{\fontsize{5.2}{5.2}\selectfont$\blacksquare$}}}
\newcommand{\sssl}{\textcolor{famssl}{\raisebox{.18ex}{\fontsize{6}{6}\selectfont$\blacktriangle$}}}
\newcommand{\scvl}{\textcolor{famcvl}{$\bullet$}}
\newcommand{\sopen}{\textcolor{famopen}{\raisebox{.18ex}{\fontsize{6}{6}\selectfont$\blacklozenge$}}}
\newcommand{\sapi}{\textcolor{famapi}{\raisebox{.12ex}{\fontsize{6.2}{6.2}\selectfont$\bigstar$}}}
\newcommand{\fsup}[1]{\textcolor{famsup}{\ssup\,#1}}
\newcommand{\fssl}[1]{\textcolor{famssl}{\sssl\,#1}}
\newcommand{\fcvl}[1]{\textcolor{famcvl}{\scvl\,#1}}
\newcommand{\fopen}[1]{\textcolor{famopen}{\sopen\,#1}}
\newcommand{\fapi}[1]{\textcolor{famapi}{\sapi\,#1}}

\title{More Accurate, Less Human: Gestalt Grouping in Vision Models}

\author{Sudhanva Manjunath Athreya\thanks{e-mail: sudhanva-manjunath.athreya@siemens.com}\\ %
        \scriptsize University of Utah\\ %
        \scriptsize Siemens AG %
\and Sai Phani Kumar Malladi\thanks{e-mail: malladi.sai-phani-kumar@siemens.com (corresponding author)}\\ %
     \scriptsize Siemens AG}

\abstract{%
Human vision organizes what it sees into wholes: same-colored points group into series,
similar marks cohere into categories, and shapes complete into recognizable objects.
These are the Gestalt operations that visualization design builds on. Whether vision models
\emph{organize} visual content this way has not been systematically tested. We introduce a
behavioral battery
that scores models against human data from prior perception studies on four grouping
tasks:
mark-color odd-one-out, color-series counting, silhouette recognition, and object
odd-one-out. We apply it to 45 models across five training families:
\fsup{supervised}, \fssl{self-supervised}, and \fcvl{contrastive vision--language}
encoders, \fopen{open-weight VLMs}, and \fapi{closed foundation models}. The battery
reveals that agreement with human responses captures aspects of perceptual organization
that conventional performance metrics fail to distinguish, with several \fapi{closed}
models exhibiting substantially lower alignment than their benchmark accuracy would
suggest. Scoring against published perception data therefore gives visualization
research a reusable yardstick, requiring no new user study, for auditing whether the
models now entering visualization pipelines organize what they see the way their human
audience does.%
}


\begin{document}

\firstsection{Introduction}
\maketitle
Vision-language models (VLMs) are increasingly described as exhibiting human-like perceptual capabilities, demonstrating impressive performance across image understanding, visual reasoning, chart interpretation, and multimodal question answering~\cite{yue2024mmmu,masry2022chartqa}.
As these systems are deployed to read charts, evaluate designs, and stand in for human viewers in visualization pipelines~\cite{zhang2023gvil}, claims of human-like perception become load-bearing and demand evidence.
High benchmark accuracy demonstrates successful task completion, but it does not establish whether models organize visual information according to the same perceptual regularities that govern human vision.

Existing evaluations primarily measure task performance, such as visualization-literacy accuracy and benchmark scores~\cite{bendeck2024,pandey2025}. While these quantify capability, they provide limited evidence of alignment with experimentally established principles of human perception. Identical answers may arise from fundamentally different perceptual organizations, making task accuracy insufficient to determine whether VLMs perceive visual structure like humans.

Cognitive psychology provides a principled basis for this question. Decades of research have established experimentally validated regularities governing how humans organize visual information, including systematic successes and errors. Gestalt psychology~\cite{wertheimer1923} introduced core principles of perceptual organization, while graphical perception research quantified their effects on visualization interpretation~\cite{cleveland1984,heer2010,mccoleman2021,demiralp2014,szafir2018}. These findings provide validated expectations against which model behavior can be systematically evaluated.

In this work, we argue that agreement with established cognitive regularities provides principled, reproducible behavioral evidence for studying cognitive grounding in vision models. It does not imply human-like internal mechanisms, but offers a measurable, mechanism-agnostic criterion for evaluating human-like perception. This extends behavioral psychophysics to visualization: whether VLMs can serve as human proxies and organize visual content like their human audience.

We demonstrate this methodology through two foundational Gestalt principles: \textbf{Closure}, the perceptual completion of incomplete structures, and \textbf{Similarity}, grouping by shared visual or semantic attributes. Together, they capture complementary mechanisms of completion and attribute-based grouping. 
Closure and Similarity are two of the foundational Gestalt grouping laws and the operations visualization most directly recruits (color assignment is a similarity manipulation; glyph reading depends on closure); together, they span complementary regimes, grouping by shared attribute and completion of incomplete structure.
We construct a behavioral benchmark of chart-based tasks grounded in published graphical perception studies, alongside natural-image transfer tasks as controls. Each task reuses or re-renders original stimuli and transforms published human data into evaluation targets, requiring no new human studies (\cref{tab:battery}).

\begin{figure*}[t]
  \centering
  \includegraphics[width=\textwidth]{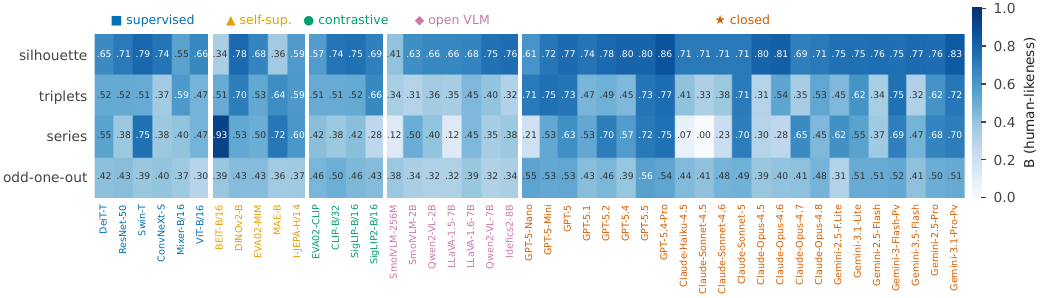}
  \caption{Human-likeness $B$ on the four battery tasks for all 45 models (tasks in rows,
  models in columns), grouped by family and ordered within each family by model size:
  parameter count for encoders and open VLMs, vendor and tier for \fapi{closed} models,
  whose sizes are undisclosed.} 
  \label{fig:matrix}
\end{figure*}

We evaluate 15 vision encoders spanning \fsup{supervised}, \fssl{self-supervised}, and \fcvl{contrastive vision--language} training, plus 30 foundation models: seven \fopen{open-weight VLMs} and 23 \fapi{closed models}. We quantify Behavioral Agreement ($B$), trial-level Behavioral Error Consistency ($\kappa$) against human--human agreement~\cite{geirhos2020,cohen1960}, and Behavioral Effect Replication ($a(k)$) across series counts. Some \fapi{closed} models achieve high accuracy and strong alignment, but many do not (\cref{fig:matrix,fig:dissociation}). Human-like grouping also fails to generalize across tasks, indicating that perceptual human-likeness is task-specific rather than a global trait.


Our contributions are threefold:
\begin{description}
\setlength{\itemsep}{1pt}
\setlength{\parskip}{0pt}
\setlength{\parskip}{0pt}
\setlength{\topsep}{0pt}
\setlength{\partopsep}{0pt}

\item[Methodology] A Gestalt-grounded behavioral evaluation for testing whether the models entering visualization pipelines organize marks and objects the way human vision does.
\item[Benchmark]  A reproducible Closure and Similarity benchmark that reuses the open-source human data.
\item[Findings] Evidence that behavioral evaluation reveals human perceptual
alignment differences missed by conventional benchmark accuracy.
\end{description}

\section{Related Work}

\paragraph{Graphical perception.}
Cleveland and McGill established the elementary-encoding hierarchy with ratio-estimation tasks~\cite{cleveland1984}, replicated at scale by Heer and Bostock~\cite{heer2010}; McColeman et al.\ revisited the ranks across channels~\cite{mccoleman2021}.
Task-level psychophysics exists for pie variants~\cite{skau2016}, bar charts~\cite{talbot2014}, categorical color in scatterplots~\cite{tseng2024,szafir2018,healey1996,ware2013}, and perceived similarity of visualization marks~\cite{demiralp2014}.
Gestalt grouping itself shapes visualization reading: spatially grouping same-colored marks speeds search and yields pop-out, with layout, mark count, and size jointly setting performance~\cite{gramazio2014}.
Several of these studies published raw per-trial data; our battery is built directly on those releases.

\paragraph{Model--human comparison.}
Behavioral comparison of deep networks and human vision has matured from accuracy gaps to error structure: texture bias~\cite{geirhos2019}, trial-level error consistency~\cite{geirhos2020,geirhos2021}, and representational alignment on human similarity judgments~\cite{muttenthaler2023,hebart2020,hebart2023}, consolidated in benchmark platforms for brain and behavioral alignment~\cite{schrimpf2020}.
Gestalt organization has itself been probed in vision models, through diagnostic disruption stimuli and rule reasoning rather than agreement with human responses~\cite{li2025gestalt,sha2025gestalt}.
In the taxonomy of model--human alignment~\cite{sucholutsky2023}, ours is the \emph{behavioral} level: scored on responses, not embeddings, so encoders and foundation models are directly comparable.
Task batteries have precedent on both sides: L-POST screens perceptual organization in humans across 15 subtests~\cite{torfs2014}, and GPT-3 was assessed on a battery of canonical cognitive experiments~\cite{binz2023}.
We import error-consistency $\kappa$, its human--human ceiling, and odd-one-out agreement into the visualization domain.

\paragraph{Models reading visualizations.}
Prior evaluations occupy two rungs.
At the level of \emph{elementary encodings}, Haehn et al.\ tested CNNs on the Cleveland--McGill ratio-estimation tasks~\cite{haehn2019}, extended to multimodal models under graphical-perception theory~\cite{zhang2025graphical}.
At the level of \emph{whole charts}, multimodal models are scored on visualization literacy and chart QA~\cite{bendeck2024,pandey2025}; CHART-6 goes furthest, correlating model and human item difficulty across six literacy instruments against a split-half human ceiling~\cite{verma2025chart6}.
The rung between them is ours.
Chart comprehension presupposes perceptual organization: before a model reads a legend it must group the marks the legend refers to.
Testing that layer calls for psychophysics rather than literacy instruments, with stimuli matched to a source experiment, trial-level scoring against published human responses, and a human--human consistency ceiling.

\section{Behavioral Evaluation Methodology}

Our objective is not simply to measure task performance, but to determine whether model behavior is consistent with well-established behavioral phenomena from cognitive psychology.
Rather than collect new human data, we build every task from a published perception study, reusing its stimuli and published human data as the evaluation target (sources in \cref{tab:battery}).
We instantiate the methodology with two foundational Gestalt principles, \textbf{Closure} and \textbf{Similarity}.

\begin{figure}[t]
  \centering
  \includegraphics[width=\columnwidth]{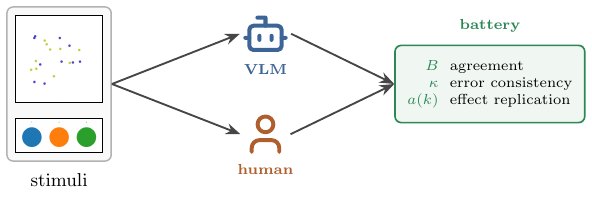}
  \caption{\textbf{The behavioral battery.} Each stimulus (left) is answered by every model and
  by human observers (center). The paired responses are then scored by three battery
  metrics (right).}
  \label{fig:battery}
\end{figure}

For each model and task, we present the study's stimuli (\cref{fig:battery}, left), reduce the model's output to one response, and compare that response to the study's human data.
The encoder's output is summarized to cosine geometry (e.g., the odd-one-out is the item outside the most-similar pair) or a frozen probe (e.g., a nearest class prototype for silhouettes, rounded ridge readouts for series counting task). In the case of foundation models, the output text is parsed by rule to the same label.
The paired responses and human targets are then scored by $B$, $\kappa$, and $a(k)$, with bootstrap intervals over stimuli.

\begin{table}[tb]
  \centering
  \caption{The battery: four grouping tasks across two Gestalt principles and two tracks.
  $n$ = test stimuli per model.}
  \label{tab:battery}
  \scriptsize\setlength{\tabcolsep}{2.5pt}
  \begin{tabular}{@{}lrl@{}}
  \toprule
  Task (track) & $n$ & Human anchor (source) \\
  \midrule
  \multicolumn{3}{@{}l}{\emph{Closure}} \\
  \;\;silhouette recognition (natural) & 80 & 1{,}600 trials, 10 subjects~\cite{geirhos2019} \\
  \addlinespace[2pt]
  \multicolumn{3}{@{}l}{\emph{Similarity}} \\
  \;\;mark-color odd-one-out (chart) & 120 & perceptual kernels, 20 workers~\cite{demiralp2014} \\
  \;\;color-series counting (chart) & 40 & derived 6--12 band~\cite{healey1996,ware2013,szafir2018} \\
  \;\;object odd-one-out (natural) & 500 & THINGS; ceiling 0.863~\cite{hebart2020,hebart2023} \\
  \bottomrule
  \end{tabular}
\end{table}

\subsection{Tasks}
\looseness=-1 \Cref{tab:battery} summarizes the battery.
Appendix \cref{sec:supp-stimuli} shows an example stimulus for each task.
\emph{Closure} is evaluated by shape-only object recognition on filled silhouettes: a 16-way forced choice in which texture, interior detail, and local appearance are removed, so recognition must rest on global shape completion.
Per-trial responses of ten observers are public~\cite{geirhos2019}; the per-image modal response is the human behavioral reference.
\emph{Similarity} is evaluated three ways: by odd-one-out over the Tableau-10 mark colors, scored against the crowdsourced perceptual kernel~\cite{demiralp2014}; by color-series counting, scored against the published 6--12 categorical-color capacity band~\cite{healey1996,ware2013,szafir2018} because no per-trial human data exists; and, on the natural track, by THINGS object odd-one-out against its published test--retest ceiling~\cite{hebart2020,hebart2023}.
The capacity band is a limit of ensemble segmentation by color~\cite{whitney2018}: series counting asks whether similarity-based grouping exhibits the same behavioral limitation as human ensemble perception, not whether models count.

\paragraph{Construction.}
Chart-track stimuli are rendered deterministically to match the source studies (the kernel study's own palette); natural-track stimuli are the source studies' images themselves.
Montage tasks present numbered, position-shuffled options.
Ground truths are derived from each study's published data release rather than its reported summaries; the kernel's minimum-dissimilarity pair leaves the third mark as the human odd-one-out, and appendix \cref{sec:supp-stimuli} details the silhouette and object targets.

\paragraph{Scope.}
\looseness=-1 The battery probes grouping at the level psychophysics does, on the building blocks of charts (individual marks and fields of points), rather than on complete charts.
This is the level at which published human perceptual data exists (kernels for isolated marks, capacity limits for point fields), and matching the source studies' stimuli exactly is what licenses error-level scoring; an axis or a legend around the dots would invalidate the human reference.
The natural track is the complementary control: silhouettes and THINGS photographs carry no chart content, so agreement there measures domain-general grouping, and differences between the tracks are themselves informative.

\subsection{Models and procedure}
We evaluate 45 models: 15 encoders, whose design axis is training objective rather than architecture, and 30 foundation models, seven \fopen{open VLMs} and 23 \fapi{closed foundation models} that answer in free text under the protocol below.
Appendix \cref{tab:models} gives the full roster with backbones and references.

\paragraph{Readouts.}
Odd-one-out tasks use a zero-shot cosine readout (the item outside the most-similar embedding pair), identical to prior odd-one-out probes~\cite{muttenthaler2023}.
Tasks without a natural zero-shot readout use minimal probes fit on a disjoint \emph{probe-train} split and applied frozen to the test split: class prototypes for silhouettes, rounded ridge regression for series counting.
Probes read encoders generously: they estimate what is linearly decodable from the representation and cannot introduce organization that is absent.
Foundation models receive the rendered stimulus with a fixed per-task prompt (options enumerated) and answer in text; answers are parsed by rule.

\paragraph{Protocol.}
\looseness=-1 Every model receives the identical human-facing stimulus image; for odd-one-out tasks that is the numbered montage the prompt refers to (encoders instead embed the component images).
Open-VLM decoding is greedy and \fapi{closed} models ran once at temperature 0 where supported, so runs are deterministic (appendix \cref{sec:supp-metrics}).
Each schema-checked trial record makes encoders and foundation models interchangeable downstream; the appendix (\crefrange{sec:supp-stimuli}{sec:supp-impl}) details stimuli, prompts, the record schema, metric derivations, per-task tables, and hardware.

\subsection{Metrics}
Three complementary measures quantify behavioral alignment.

\paragraph{Behavioral Agreement.}
For a task with test stimuli $i=1,\dots,n$, model response $r_i$, ground truth $g_i$, and human target $h_i$, Behavioral Agreement is the human-agreement rate
\begin{equation}
  B \;=\; \frac{1}{n}\sum_{i=1}^{n} \mathbf{1}\!\left[\,r_i \simeq h_i\,\right]\;\in\;[0,1],
  \label{eq:B}
\end{equation}
\looseness=-1 with the match relation $\simeq$ instantiated per task.
For mark-color and object odd-one-out, $h_i$ \emph{is} the human choice (the perceptual-kernel pick; the THINGS modal choice) and $\simeq$ is equality.
For silhouettes, $h_i$ is the modal response of the ten observers on image $i$; accuracy $\frac{1}{n}\sum_i \mathbf{1}[r_i=g_i]$ against the true class is reported separately; the two dissociate (\cref{fig:dissociation}), which is exactly the sense in which $B$ measures human-likeness, not skill.
Series counting reports accuracy against the rendered count; its human anchor is the capacity band.
CIs bootstrap stimuli (1{,}000 resamples, fixed seed).

\paragraph{Behavioral Error Consistency.}
With model correctness $m_i=\mathbf{1}[r_i=g_i]$ and per-item human accuracy $p_i$ (the fraction of the ten observers correct on item $i$),
\begin{equation}
  \kappa \;=\; \frac{c_{\mathrm{obs}}-c_{\mathrm{exp}}}{1-c_{\mathrm{exp}}},
  \qquad
  c_{\mathrm{obs}} \;=\; \frac{1}{n}\sum_{i=1}^{n}\big[m_i p_i + (1-m_i)(1-p_i)\big],
  \label{eq:kappa}
\end{equation}
where $c_{\mathrm{exp}}=\bar m\bar p+(1-\bar m)(1-\bar p)$ is the agreement expected of independent observers at the same accuracies $\bar m,\bar p$~\cite{geirhos2020}; $\kappa>0$ means the model errs on the same \emph{items} people do, beyond what accuracy alone forces.
Its ceiling is the mean pairwise $\kappa$ of the ten observers.
Appendix \cref{sec:supp-metrics} derives both estimators, their guessing floors, and the ceiling's invariance.

\paragraph{Behavioral Effect Replication.}
Signature human effects should reappear in model behavior.
The battery reads $a(k)$, accuracy as a function of the rendered series count $k$, against the human capacity band: replication means the profile bends where human capacity does.

\section{Behavioral Evidence of Cognitive Grounding}

\begin{figure}[t]
  \centering
  \includegraphics[width=\columnwidth]{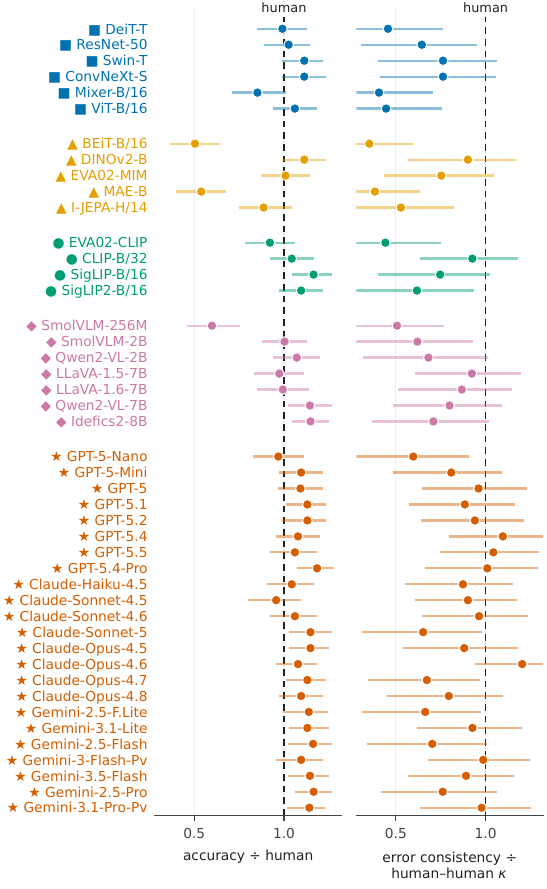}
  \caption{Accuracy and human-likeness are different properties. One row per model,
  grouped by family (colors and symbols as in \cref{fig:matrix}) and ordered within each
  family by model size as in \cref{fig:matrix}; lines are bootstrapped 95\% CIs, dashed
  lines the human references. }
  \label{fig:dissociation}
\end{figure}

\paragraph{How results are read.}
\looseness=-1 Each score is compared against source-study anchors, not chosen by us: position-guessing floors for the three-alternative tasks, chance for encoders, human references, and the categorical-color capacity band (see Appendix \cref{sec:supp-metrics}). Values are reported as point estimates, with bootstrap 95\% CIs where claims depend on them; \cref{fig:dissociation} shows every interval, and released scores tabulate them.

\paragraph{Behavioral agreement reveals properties beyond conventional accuracy}

\looseness=-1 Behavioral agreement is multi-dimensional rather than a single property (\cref{fig:matrix}): across the battery, model rankings agree only weakly, with cross-task rank correlations spanning $\rho=0.15$ to $0.62$.
The leaders trade places: the best color-similarity encoder (DINOv2) sits mid-pack on semantic odd-one-out, and the best semantic encoder (CLIP, consistent with prior THINGS probes~\cite{muttenthaler2023,hebart2020}) is unremarkable on color.
Rank reversals like these recur in every family block: which model ``looks aligned'' depends on which grouping task one measures.

Accuracy then dissociates from human-likeness on identical trials (\cref{fig:dissociation}).
Silhouettes are the battery's dissociation instrument, the one task where a model receives both an accuracy against the class labels and a $\kappa$ against the ten observers' error pattern; series counting, which has no per-trial human data, is read against the coarser capacity band and so carries no $\kappa$.
Thirty-four of the 45 models answer \emph{more accurately than the human observers}; four of them match human--human error consistency.
The two measures rank models discordantly in 37\% of pairs ($\rho=0.38$), so a model's accuracy says little about whether it errs where people do.
\fcvl{Contrastive vision--language} encoders show both directions at once: SigLIP passes human accuracy while sharing little of the human error pattern, whereas CLIP nearly reaches the human--human ceiling at human-level accuracy.
Among \fopen{open VLMs} the ordering inverts outright: every one that exceeds human accuracy falls below CLIP-tower \fopen{LLaVA-1.5-7B}, which never reaches human accuracy yet holds the tier's highest consistency.

Only at the frontier do the two axes meet.
Four \fapi{closed} models pair above-human accuracy with $\kappa$ at or beyond the ceiling in point estimate, led by Claude-Opus-4.6, whose interval contains the ceiling.
This is parity with human--human consistency rather than superiority.
Their tier does not follow them: of the 34 models above human accuracy, 30 stay below the consistency ceiling, at a median 78\% of it.
Several are tier-mates that match the leaders' accuracy.
Nor is the dissociation a measurement artifact: $\kappa$ is a chance-corrected covariance that accuracy leaves free (appendix \cref{sec:supp-metrics}).
Accuracy does not purchase human-likeness at any scale, but at the frontier, some training recipes buy both at once.

\paragraph{Grouping is a model-family property}

\looseness=-1 Whether grouping is expressed at all is a family property, and the within-family ordering of \cref{fig:matrix} makes it visible.
The seven \fopen{open VLMs} (256M--8B) sit at the bottom of both kernels: none clears the position-guessing floor on semantic odd-one-out, and only three clear it on the color kernel, the best of them by 0.10.
Little of the organization their vision towers carry reaches their generated answers.
The 23 \fapi{closed foundation models} invert this, and do so as a block rather than through a few outliers: ten exceed the best encoder on semantic grouping and eight on the color kernel.
Pairwise intervals overlap, so no single comparison carries the claim; the shift of the whole distribution does.
Neither answering format nor raw scale is the limit, since the tier's smallest model (GPT-5-Nano) clears both encoder bests.

\paragraph{Training objectives determine organization}

Which grouping becomes human-like tracks the training objective, not the model class.
Among encoders, \fssl{self-supervised} training leads the color kernel and \fcvl{contrastive vision--language} leads semantic grouping and shape, while masked-image training holds the two lowest shape consistencies (BEiT, MAE).
Behavioral Effect Replication fails in every family: series scores range more than threefold across encoder objectives, with $a(k)$ unconstrained by the human 6--12 band.
The same spread recurs in the \fapi{closed} tier: each vendor line spans at least 0.32 on the color kernel, as wide as the whole encoder range, so tier-mates differ as much as differently trained encoders.
Human-alignment claims should therefore be indexed by task and training family, not per model class.

\section{Implications for Visualization}          
\label{sec:discussion}

\paragraph{What behavioral evaluation establishes.} 

The methodology generalizes past the models scored here and asks whether their behavior agrees with experimentally established regularities of human perception and not whether it can solve perceptual tasks. We treat behavioral agreement as initial evidence of cognitive grounding, a distinction the battery earns empirically, since the models that benchmark alike diverge significantly behaviorally. The battery brings the tools of psychophysics, such as error-consistency, human–human ceiling, and odd-one-out agreement into VLM evaluation.

\paragraph{Vision models as human proxies.}
\looseness=-1 
Alignment is task-dependent (\cref{fig:matrix}): a model that matches human grouping on one task can diverge sharply on another, so no single accuracy score certifies a model as a stand-in for a human viewer.
This cautions against treating vision models as drop-in \emph{human proxies} for design evaluation or assuming that human perceptual guidelines automatically transfer to machine readers: their grouping of chart marks can depart substantially from human behavior.
These uses are already emerging in visualization pipelines, where vision models caption charts, recommend encodings, generate alt text, and screen designs on a human's behalf; the audit says which of these a given model can be trusted to do.
What it calls for instead is a task-indexed behavioral benchmark: error-consistency against published perception data is a cheap, reusable yardstick, needing no new user study, for auditing any model before it is trusted to ``see'' charts the way people do.
As the audits are task-indexed, it extends upward from isolated marks and objects to complete charts, where context effects (axes, legends, layout) can be audited task by task, turning our battery into a full-stack check on the model being used to read the visualizations.

\section{Conclusion}
\looseness=-1 Treating published psychophysics as reusable ground truth, we introduced a theory-driven behavioral battery to test whether the visual representations behind modern chart readers group marks and objects the way human vision does.
Mostly they do not: accuracy and human-likeness dissociate across every family, and \fopen{open VLMs} express little of the grouping their encoder relatives show.
Yet the \fapi{closed foundation models} show this ceiling is not principled, their best matching or exceeding human--human consistency on shape while their tier-mates still dissociate.
Behavioral grounding in human vision is measurable, task-indexed, and, on current evidence, earned by training recipe, not by accuracy or scale.

\paragraph{Limitations.}
Family and scale covary in the open tier, so the scale control comes from the \fapi{closed} tier, where the dissociation appears at every scale tested (\cref{fig:dissociation}).
Residual nulls are excluded rather than penalized, stay under 3\% per \fapi{closed-model} battery, and are dominated by content-moderation refusals on a few natural images (per-task counts: appendix \cref{tab:supp-parse}).
No stimulus is a complete chart (\S3): chart-context effects (axes, legends, occlusion, data semantics) remain untested here.
The battery covers two Gestalt principles; the remaining grouping laws are outside its scope.

\paragraph{Future directions.}
The methodology is general: the same construction, a published study supplying stimuli and human behavioral anchors, applies to any cognitive theory with experimentally established behavioral regularities.
Closure and Similarity are the first demonstration because both possess high-quality published human data; extending the battery to proximity, continuity, and the remaining grouping laws, and to broader theories of perception and cognition, is the immediate next step, with each addition inheriting the same anchors and scoring.

\acknowledgments{%
We thank Paul Rosen of the University of Utah for his reviews of the manuscript and for
feedback that sharpened its framing and presentation. We also thank Sharmila S.\ M.\ of
Siemens AG for supporting this work and helping coordinate the project.%
}

\clearpage
\bibliographystyle{abbrv-doi}
\bibliography{refs}

\clearpage
\setcounter{section}{0}\setcounter{table}{0}\setcounter{figure}{0}\setcounter{equation}{0}
\renewcommand{\thesection}{\Alph{section}}
\renewcommand{\thetable}{A\arabic{table}}
\renewcommand{\thefigure}{A\arabic{figure}}
\renewcommand{\theequation}{A\arabic{equation}}
\renewcommand{\theHsection}{apx\Alph{section}}   
\renewcommand{\theHtable}{A\arabic{table}}
\renewcommand{\theHfigure}{A\arabic{figure}}
\renewcommand{\theHequation}{A\arabic{equation}}
\section*{Appendix}
\section{Example stimuli}
\label{sec:supp-stimuli}

Every stimulus below is the exact image presented to the foundation models (for
odd-one-out tasks, the numbered montage the prompt refers to); encoders receive the
component images for the montage tasks (the three sources) and the stimulus itself
otherwise. One example per task, drawn from the test split.

\begin{figure}[htb]
  \centering
  \includegraphics[width=0.9\columnwidth]{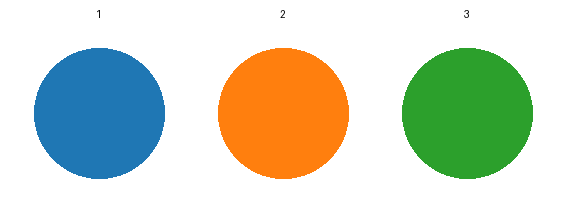}
  \caption{\textbf{Mark-color odd-one-out} (similarity, chart track): three Tableau-10 mark
  colors as numbered swatches (stimulus \texttt{mt\_000\_012}); the target is the
  crowdsourced perceptual-kernel choice~\cite{demiralp2014} (here, mark 2). Swatches are
  synthesized from the palette hexes so no digit or shape cue leaks into the embedding.}
  \label{fig:supp-triplets}
\end{figure}

\begin{figure}[htb]
  \centering
  \includegraphics[width=0.46\columnwidth]{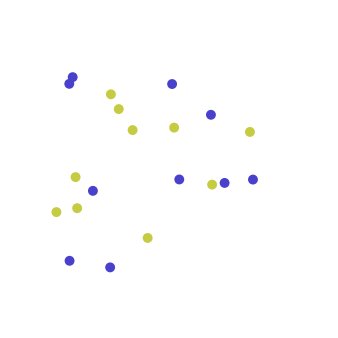}
  \caption{\textbf{Series counting} (similarity, chart track): how many distinct color
  series does the scatterplot contain (stimulus \texttt{cs\_k02\_p0}; $K=2$)? Scored
  against the rendered count, with the human 6--12 capacity band as the effect
  reference~\cite{healey1996,ware2013,szafir2018}.}
  \label{fig:supp-series}
\end{figure}

\begin{figure}[htb]
  \centering
  \includegraphics[width=0.38\columnwidth]{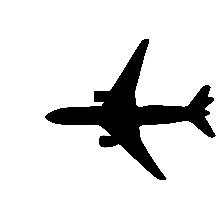}
  \caption{\textbf{Silhouette recognition} (closure, natural track): 16-way forced choice on
  filled silhouettes (stimulus \texttt{silh\_airplane\_0}); per-trial responses of ten human
  observers are public~\cite{geirhos2019}, enabling modal-match $B$ and error-consistency
  $\kappa$.}
  \label{fig:supp-silhouette}
\end{figure}

\begin{figure}[htb]
  \centering
  \includegraphics[width=\columnwidth]{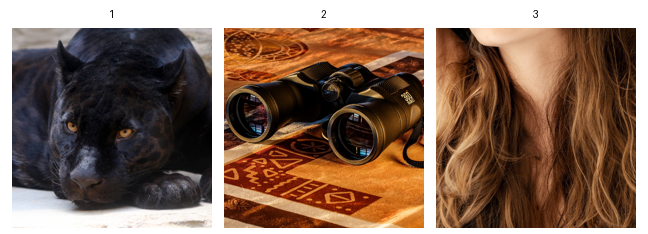}
  \caption{\textbf{Object odd-one-out} (similarity, natural track): THINGS triplet
  (stimulus \texttt{ooo\_0000}); the target is the human modal choice (here, object 2), with
  the 0.863 test--retest ceiling~\cite{hebart2020,hebart2023}.}
  \label{fig:supp-ooo}
\end{figure}

\section{Verbatim prompts}
\label{sec:supp-prompts}

One fixed prompt per task, fully literal; the image is attached before the text.

\begin{itemize}\setlength\itemsep{2pt}\scriptsize\raggedright
\item \textbf{series counting:} \texttt{How many distinct color series (groups of
same-colored points) are in this scatterplot? Answer with a single integer.}
\item \textbf{mark-color odd-one-out:} \texttt{Three colored chart marks are shown, numbered
1, 2 and 3. Which one is the odd one out (least similar in color to the other two)? Answer
1, 2, or 3.}
\item \textbf{object odd-one-out:} \texttt{Three objects are shown, numbered 1, 2 and 3.
Which one is the odd one out (least similar to the other two)? Answer 1, 2, or 3.}
\item \textbf{silhouette:} \texttt{What object is shown in this black silhouette? Choose
exactly one: airplane, bear, bicycle, bird, boat, bottle, car, cat, chair, clock, dog,
elephant, keyboard, knife, oven, truck.}
\end{itemize}

Answers are parsed by rule: for choices and counts, the first integer in the response
(spelled-out number words accepted, e.g.\ ``One.''\ $\rightarrow$ 1), constrained to the
option set; for text-label choices, the first option mentioned. Anything else is recorded
as \texttt{null}; an
unparseable answer is never scored as an error.

\section{Per-trial record schema}
\label{sec:supp-schema}

Every model $\times$ stimulus trial emits one JSON record; scoring consumes only these
records, so encoders and foundation models are interchangeable downstream. A verbatim
example
(\texttt{raw\_output} truncated):

{\tiny\verbatiminput{tables/record_example.json}}

\section{Metric derivations and rationale}
\label{sec:supp-metrics}

The main paper states the two headline statistics compactly (its Eqs.~1--2); this section
derives them, fixes their floors and ceilings exactly, and records the reasoning behind
each scoring decision. Throughout, $i=1,\dots,n$ indexes test stimuli, $r_i$ is the model
response, $g_i$ the ground truth, $h_i$ the human target, and
$m_i=\mathbf{1}[r_i=g_i]$ model correctness.

\subsection{What Behavioral Agreement $B$ estimates, and why it is not accuracy}

$B=\frac1n\sum_i\mathbf{1}[r_i\simeq h_i]$ is the plug-in estimate of $\Pr(r\simeq h)$:
the probability that the model reproduces the human reference behavior on a stimulus drawn
from the task's stimulus distribution. Each summand is a Bernoulli indicator, so
$\mathrm{Var}(B)=p(1-p)/n\le 1/(4n)$ and the 95\% CI half-width is at most
$1.96/(2\sqrt n)$: 0.110 for silhouettes ($n{=}80$), 0.089 for mark triplets (120), 0.155
for series counting (40), and 0.044 for object odd-one-out (500). These bounds set how
each battery may be read: the 500-stimulus THINGS sample supports the frontier-tier
comparisons, where neighboring models differ by a few points; the 80-image silhouette test
split (the half of the Geirhos set not used for probe training) is read through $\kappa$
and the accuracy--$B$ dissociation rather than $B$ rank order; and 40-stimulus series
counting is read only as capacity-band membership.

Accuracy and $B$ separate only where the human target disagrees with the ground truth: on
items with $h_i=g_i$ the indicators $\mathbf{1}[r_i=g_i]$ and $\mathbf{1}[r_i=h_i]$
coincide, so
\begin{equation}
  \mathrm{acc}-B \;=\; \frac{1}{n}\sum_{i:\,h_i\neq g_i}
  \big(\mathbf{1}[r_i=g_i]-\mathbf{1}[r_i=h_i]\big):
  \label{eq:acc-b}
\end{equation}
accuracy exceeds $B$ exactly by how often the model is right where the typical human is
wrong, minus how often it commits the very error the humans commit. A model therefore
cannot raise $B$ by being superhumanly good (solving human-hard items moves accuracy up
and $B$ down), which is the sense in which $B$ measures behavioral coincidence rather
than skill. For the two odd-one-out tasks the distinction collapses by construction
($g_i$ \emph{is} the human choice, so $B$ is accuracy against it); silhouettes, with
class labels independent of the ten observers' per-trial responses, are the one task where
both quantities exist separately, which is what makes them the battery's dissociation
instrument (main-paper Fig.~3).

\subsection{Behavioral Error Consistency: the chance correction, and both sides of the ceiling}

Raw response overlap is the wrong consistency measure because it is dominated by the
marginals: two observers answering \emph{independently} with accuracies $\bar m,\bar p$
still agree at rate $c_{\mathrm{exp}}=\bar m\bar p+(1-\bar m)(1-\bar p)$: 0.905 at
$\bar m=\bar p=0.95$, and already 0.60 at our silhouette panel's mean accuracy (0.719).
Overlap therefore mostly restates accuracy. Cohen's correction~\cite{cohen1960}, applied
to correctness vectors following~\cite{geirhos2020}, rescales the excess over independence
to the attainable range,
\begin{equation}
  \kappa \;=\; \frac{c_{\mathrm{obs}}-c_{\mathrm{exp}}}{1-c_{\mathrm{exp}}},
  \label{eq:supp-kappa}
\end{equation}
so $\kappa=0$ holds exactly when errors fall on items independently given the accuracies:
$\kappa$ isolates \emph{which items} a model fails, with the accuracy contribution
removed.

An equivalent form makes the geometry explicit. Expanding both agreement terms,
$c_{\mathrm{obs}}-c_{\mathrm{exp}}
=2\big[\frac1n\sum_i m_i p_i-\bar m\bar p\,\big]=2\,\mathrm{Cov}(m,p)$, so
\begin{equation}
  \kappa \;=\; \frac{2\,\mathrm{Cov}(m,p)}{1-c_{\mathrm{exp}}}:
  \label{eq:supp-kappa-cov}
\end{equation}
error consistency is exactly a rescaled covariance between model correctness and
per-item human accuracy, i.e.\ a \emph{difficulty-alignment} measure. Accuracy constrains
only the mean $\bar m$; $\kappa$ is a second moment, whether the model finds hard what
people find hard, and the two are free to vary independently, which is why the two
columns of main-paper Fig.~3 can disagree so thoroughly. In correlation units, SigLIP's
above-human accuracy comes with $r(m,p)=0.53$ while Claude-Opus-4.6 reaches $r=0.81$;
both identities hold to machine precision on our data (\texttt{scoring.supp\_stats}).

Our estimator differs from the pairwise original in two conventions, and both are
validated on this panel. \emph{(i) Reference.} Geirhos et al.\ score a model against each
observer separately and average the ten $\kappa$'s; main-paper Eq.~2 scores once against
the panel's per-item accuracy $p_i$. The observed-agreement terms are identical by
linearity, $\frac1{10}\sum_j c_{\mathrm{obs}}(m,h_j)=c_{\mathrm{obs}}(m,p)$, so the two
estimators differ only through the chance correction's nonlinearity across observers of
different accuracy (panel range 0.59--0.84): across all 45 paper models the gap is at most
0.013. We use the aggregate form because it has one reference, one number, and a matching
stimulus bootstrap. \emph{(ii) Ceiling.} The 0.489 ceiling is the mean over the 45
observer pairs (pairwise $\kappa$ range 0.150--0.812; individual humans differ widely),
while a model is scored against the ten-observer mean, and matching a mean could in
principle be easier than matching one noisy individual. Empirically the asymmetry is nil:
scoring each observer against the leave-one-out mean of the other nine gives mean
$\kappa=0.487$ (range 0.313--0.616), indistinguishable from the pairwise 0.489. The
ceiling is convention-invariant here, and the strongest frontier values
($\kappa=0.54$--$0.59$) read as \emph{above the average human pair yet within the observer
distribution's upper range} (best single observer vs.\ panel: 0.616): as
panel-consistent as the most panel-typical humans, not superhumanly consistent.

\subsection{Floors and ceilings: guessing is better than 1/3}

For a three-alternative task the uniform-guess baseline is $1/3$, but the achievable
zero-perception floor is higher: a model that always answers position $k$ scores the
empirical target frequency $f_k=\frac1n\sum_i\mathbf{1}[h_i=k]$, so the floor is
$\max_k f_k$: 0.35 for mark triplets ($f=0.31/0.34/0.35$) and 0.39 for object
odd-one-out ($f=0.39/0.31/0.30$). Several small open VLMs emit a near-constant position
and land exactly there, which is why the main text reads scores $\le$0.35/0.39 as
``position-guessing level'' rather than comparing them with $1/3$. Silhouette classes are
exactly balanced by construction ($16\times5$ test images), so its floor is the uniform
$1/16\approx0.06$ with no majority-class shortcut; $\kappa$'s floor is 0 by construction,
the chance correction subtracting precisely the guessing component. On the other side,
odd-one-out $B$ is bounded in practice by the 0.863 test--retest consistency of the human
modal choice itself (matching a consensus more often than humans re-match it is not
meaningful), so the main text quotes those scores as fractions of that ceiling where
rank claims are made.

\subsection{Aggregation, determinism, and nulls}

Open-VLM decoding is greedy, capped at 40 new tokens, with one repetition per stimulus,
so locally run numbers are deterministic given the stimuli; closed models ran once at
temperature 0 where supported, with output budgets sized for thinking tiers; the aggregation rule (majority vote over repetitions for
choice and count tasks, mean for scalar ones) exists for reproduction runs at nonzero
temperature and is the identity here. Unparseable answers are recorded as null and
\emph{excluded}, never scored as errors: a refusal or a format break is evidence about
instruction-following, not perceptual organization. Exclusion changes the estimand to
$\Pr(r\simeq h\mid\mathrm{parseable})$; with null fraction $\nu$ the fully-scored value is
bracketed by $(1-\nu)B\le B_{\mathrm{full}}\le(1-\nu)B+\nu$, a displacement of at most
$\nu$. The worst per-task null rate in the paper is 7.5\% (six of 80 silhouettes;
\cref{tab:supp-parse}), below that task's maximal CI half-width of 0.110, and 16 of the 30
foundation-model batteries are null-free throughout (encoders always respond), so no conclusion depends
on this convention.

\subsection{Bootstrap design}

Stimuli are the only sampled quantity in the design (model responses are deterministic
and the human references fixed), so all intervals resample stimuli with replacement and
take percentile bounds: 1{,}000 draws (fixed seed) for $B$, 2{,}000 for the $\kappa$
intervals of main-paper Fig.~3, recomputing the full statistic (both $c_{\mathrm{obs}}$
and $c_{\mathrm{exp}}$) in every draw so the interval carries the chance correction's own
sampling noise. Because every model answers the identical stimulus set, between-model
contrasts are paired, and the marginal CIs drawn in the figures are conservative for
model-vs-model comparisons; a paired bootstrap of the difference would be tighter than
the overlap of two marginal intervals.

\section{Implementation and hardware}
\label{sec:supp-impl}

The battery regenerates end to end: sources auto-download (the perceptual-kernels
release, the raw Geirhos trials, the THINGS triplets and images), every build step is
seeded, and all stimuli, human baselines, per-trial records, and scores reproduce
deterministically from public sources. Each trial emits one schema-checked record (the
schema above) carrying the model version, prompt hash, and run id, so every number in
the paper traces to its run.

Local runs (encoders and open VLMs through 8B) executed on a single laptop:
\begin{itemize}\setlength\itemsep{1pt}
  \item CPU: Intel Core Ultra 9 185H
  \item GPU: NVIDIA GeForce RTX 4090 Laptop GPU (16\,GB)
  \item RAM: 32\,GB
\end{itemize}
Qwen2-VL-7B ran through the same harness on a Lambda Vector One workstation:
\begin{itemize}\setlength\itemsep{1pt}
  \item CPU: AMD Ryzen Threadripper 7960X (48 threads)
  \item GPU: 2$\times$ NVIDIA GeForce RTX 4090
  \item RAM: 256\,GB DDR5
\end{itemize}
Precision does not move the numbers: LLaVA-1.5, run at fp16 on the laptop and at full
precision on the workstation, reproduces its mark-triplet score and error consistency
exactly (odd-one-out within 4pp). The closed models are served remotely, so their batteries required no local compute
beyond stimulus upload.

\section{Extended results}
\label{sec:supp-results}

\begin{table}[htb]
  \centering
  \caption{The 45 evaluated models, by family. Encoders differ by training objective on
  comparable backbones (DeiT-T to I-JEPA-H/14; most ViT-B); the foundation models answer
  every task as free text.}
  \label{tab:models}
  \scriptsize\setlength{\tabcolsep}{3pt}
  \begin{tabular}{@{}lp{0.60\columnwidth}@{}}
  \toprule
  family & models \\
  \midrule
  \fsup{supervised} (6) & ViT-B/16, ResNet-50, DeiT-T, Swin-T, ConvNeXt-S, MLP-Mixer-B/16 \\
  \fssl{self-supervised} (5) & BEiT-B/16, MAE-B~\cite{he2022}, DINOv2-B~\cite{oquab2023},
    I-JEPA-H/14, EVA02-MIM \\
  \fcvl{contrastive V--L} (4) & CLIP-B/32~\cite{radford2021}, SigLIP-B/16~\cite{zhai2023},
    SigLIP2-B/16, EVA02-CLIP \\
  \fopen{open VLM} (7) & SmolVLM-256M, SmolVLM-2B~\cite{marafioti2025smolvlm},
    Qwen2-VL-2B and -7B~\cite{wang2024qwen2vl}, LLaVA-1.5-7B~\cite{liu2024llava15},
    LLaVA-1.6-Mistral-7B, Idefics2-8B \\
  \fapi{closed} (23) & GPT-5 (base, Mini, Nano, 5.1, 5.2, 5.4, 5.4-Pro, 5.5); Claude
    (Haiku-4.5, Sonnet-4.5/4.6/5, Opus-4.5/4.6/4.7/4.8); Gemini (2.5-Flash/-Lite/-Pro,
    3-Flash-Preview, 3.1-Flash-Lite, 3.1-Pro-Preview, 3.5-Flash) \\
  \bottomrule
  \end{tabular}
\end{table}

\begin{table}[htb]
  \centering\scriptsize\setlength{\tabcolsep}{3pt}
  \caption{Human-likeness $B$ on the four paper tasks for all 45 evaluated models,
  grouped by training family (symbols and colors as in the main paper: \fsup{supervised},
  \fssl{self-supervised}, \fcvl{contrastive V--L}, \fopen{open VLM}, \fapi{closed}).}
  \label{tab:supp-matrix}
  \begin{tabular}{@{}lrrrr@{}}
\toprule
model & silh. & trip. & series & ooo \\
\midrule
\ssup~ViT-B/16 & 0.662 & 0.475 & 0.475 & 0.296 \\
\ssup~ResNet-50 & 0.713 & 0.517 & 0.375 & 0.426 \\
\ssup~DeiT-T & 0.650 & 0.517 & 0.550 & 0.416 \\
\ssup~Swin-T & 0.787 & 0.508 & 0.750 & 0.390 \\
\ssup~ConvNeXt-S & 0.738 & 0.367 & 0.375 & 0.402 \\
\ssup~Mixer-B/16 & 0.550 & 0.592 & 0.400 & 0.374 \\
\midrule
\sssl~BEiT-B/16 & 0.338 & 0.508 & 0.925 & 0.390 \\
\sssl~MAE-B & 0.362 & 0.642 & 0.725 & 0.364 \\
\sssl~DINOv2-B & 0.775 & 0.700 & 0.525 & 0.432 \\
\sssl~I-JEPA-H/14 & 0.588 & 0.592 & 0.600 & 0.370 \\
\sssl~EVA02-MIM & 0.675 & 0.525 & 0.500 & 0.434 \\
\midrule
\scvl~CLIP-B/32 & 0.738 & 0.508 & 0.375 & 0.498 \\
\scvl~SigLIP-B/16 & 0.750 & 0.517 & 0.425 & 0.456 \\
\scvl~SigLIP2-B/16 & 0.688 & 0.658 & 0.275 & 0.434 \\
\scvl~EVA02-CLIP & 0.575 & 0.508 & 0.425 & 0.462 \\
\midrule
\sopen~SmolVLM-256M & 0.405 & 0.342 & 0.125 & 0.382 \\
\sopen~SmolVLM-2B & 0.633 & 0.308 & 0.500 & 0.336 \\
\sopen~Qwen2-VL-2B & 0.662 & 0.358 & 0.400 & 0.322 \\
\sopen~LLaVA-1.5-7B & 0.662 & 0.350 & 0.125 & 0.322 \\
\sopen~LLaVA-1.6-7B & 0.675 & 0.450 & 0.450 & 0.386 \\
\sopen~Qwen2-VL-7B & 0.747 & 0.400 & 0.350 & 0.317 \\
\sopen~Idefics2-8B & 0.762 & 0.317 & 0.375 & 0.336 \\
\midrule
\sapi~GPT-5 & 0.772 & 0.733 & 0.632 & 0.531 \\
\sapi~GPT-5-Mini & 0.725 & 0.750 & 0.525 & 0.526 \\
\sapi~GPT-5-Nano & 0.608 & 0.714 & 0.210 & 0.546 \\
\sapi~GPT-5.1 & 0.738 & 0.467 & 0.525 & 0.428 \\
\sapi~GPT-5.2 & 0.775 & 0.492 & 0.700 & 0.460 \\
\sapi~GPT-5.4 & 0.800 & 0.450 & 0.575 & 0.392 \\
\sapi~GPT-5.4-Pro & 0.865 & 0.767 & 0.750 & 0.538 \\
\sapi~GPT-5.5 & 0.800 & 0.735 & 0.718 & 0.562 \\
\sapi~Claude-Haiku-4.5 & 0.713 & 0.408 & 0.075 & 0.444 \\
\sapi~Claude-Sonnet-4.5 & 0.713 & 0.325 & 0.000 & 0.412 \\
\sapi~Claude-Sonnet-4.6 & 0.713 & 0.375 & 0.225 & 0.476 \\
\sapi~Claude-Sonnet-5 & 0.713 & 0.708 & 0.700 & 0.488 \\
\sapi~Claude-Opus-4.5 & 0.800 & 0.308 & 0.300 & 0.388 \\
\sapi~Claude-Opus-4.6 & 0.812 & 0.542 & 0.275 & 0.404 \\
\sapi~Claude-Opus-4.7 & 0.688 & 0.350 & 0.650 & 0.412 \\
\sapi~Claude-Opus-4.8 & 0.713 & 0.533 & 0.450 & 0.476 \\
\sapi~Gemini-2.5-Flash & 0.759 & 0.342 & 0.368 & 0.509 \\
\sapi~Gemini-2.5-F.Lite & 0.753 & 0.450 & 0.625 & 0.312 \\
\sapi~Gemini-2.5-Pro & 0.762 & 0.617 & 0.675 & 0.502 \\
\sapi~Gemini-3-Flash-Pv & 0.750 & 0.750 & 0.692 & 0.524 \\
\sapi~Gemini-3.1-Lite & 0.750 & 0.617 & 0.550 & 0.508 \\
\sapi~Gemini-3.1-Pro-Pv & 0.833 & 0.725 & 0.700 & 0.514 \\
\sapi~Gemini-3.5-Flash & 0.772 & 0.319 & 0.475 & 0.413 \\
\bottomrule
\end{tabular}

\end{table}

\begin{table}[htb]
  \centering\scriptsize\setlength{\tabcolsep}{3pt}
  \caption{Silhouette detail per model: class accuracy with bootstrapped 95\% CI,
  error-consistency $\kappa$ against the ten-observer panel, and modal-match $B$.}
  \label{tab:supp-silhouette}
  \begin{tabular}{@{}lrrr@{}}
\toprule
model & acc.\ [95\% CI] & $\kappa$ & $B$ (modal) \\
\midrule
humans & 0.719 & 0.489 (ceiling) & --- \\
\midrule
\ssup~ViT-B/16 & 0.762 [0.68,0.85] & 0.218 & 0.662 \\
\ssup~ResNet-50 & 0.738 [0.64,0.82] & 0.315 & 0.713 \\
\ssup~DeiT-T & 0.713 [0.61,0.81] & 0.223 & 0.650 \\
\ssup~Swin-T & 0.800 [0.71,0.88] & 0.373 & 0.787 \\
\ssup~ConvNeXt-S & 0.800 [0.71,0.89] & 0.373 & 0.738 \\
\ssup~Mixer-B/16 & 0.613 [0.51,0.72] & 0.199 & 0.550 \\
\midrule
\sssl~BEiT-B/16 & 0.362 [0.26,0.46] & 0.172 & 0.338 \\
\sssl~MAE-B & 0.388 [0.29,0.49] & 0.188 & 0.362 \\
\sssl~DINOv2-B & 0.800 [0.71,0.89] & 0.441 & 0.775 \\
\sssl~I-JEPA-H/14 & 0.637 [0.54,0.75] & 0.258 & 0.588 \\
\sssl~EVA02-MIM & 0.725 [0.62,0.82] & 0.368 & 0.675 \\
\midrule
\scvl~CLIP-B/32 & 0.750 [0.66,0.84] & 0.453 & 0.738 \\
\scvl~SigLIP-B/16 & 0.838 [0.75,0.91] & 0.365 & 0.750 \\
\scvl~SigLIP2-B/16 & 0.787 [0.70,0.88] & 0.302 & 0.688 \\
\scvl~EVA02-CLIP & 0.662 [0.56,0.76] & 0.216 & 0.575 \\
\midrule
\sopen~SmolVLM-256M & 0.430 [0.33,0.54] & 0.248 & 0.405 \\
\sopen~SmolVLM-2B & 0.722 [0.63,0.81] & 0.303 & 0.633 \\
\sopen~Qwen2-VL-2B & 0.770 [0.68,0.86] & 0.333 & 0.662 \\
\sopen~LLaVA-1.5-7B & 0.700 [0.60,0.80] & 0.452 & 0.662 \\
\sopen~LLaVA-1.6-7B & 0.714 [0.61,0.82] & 0.424 & 0.675 \\
\sopen~Qwen2-VL-7B & 0.823 [0.73,0.91] & 0.391 & 0.747 \\
\sopen~Idefics2-8B & 0.825 [0.75,0.90] & 0.347 & 0.762 \\
\midrule
\sapi~GPT-5 & 0.785 [0.70,0.87] & 0.470 & 0.772 \\
\sapi~GPT-5-Mini & 0.787 [0.70,0.88] & 0.395 & 0.725 \\
\sapi~GPT-5-Nano & 0.696 [0.59,0.80] & 0.292 & 0.608 \\
\sapi~GPT-5.1 & 0.812 [0.72,0.89] & 0.432 & 0.738 \\
\sapi~GPT-5.2 & 0.812 [0.71,0.89] & 0.460 & 0.775 \\
\sapi~GPT-5.4 & 0.775 [0.69,0.86] & 0.536 & 0.800 \\
\sapi~GPT-5.4-Pro & 0.851 [0.77,0.92] & 0.493 & 0.865 \\
\sapi~GPT-5.5 & 0.762 [0.66,0.85] & 0.510 & 0.800 \\
\sapi~Claude-Haiku-4.5 & 0.750 [0.65,0.84] & 0.427 & 0.713 \\
\sapi~Claude-Sonnet-4.5 & 0.688 [0.57,0.79] & 0.441 & 0.713 \\
\sapi~Claude-Sonnet-4.6 & 0.762 [0.66,0.85] & 0.471 & 0.713 \\
\sapi~Claude-Sonnet-5 & 0.825 [0.74,0.91] & 0.319 & 0.713 \\
\sapi~Claude-Opus-4.5 & 0.825 [0.74,0.90] & 0.431 & 0.800 \\
\sapi~Claude-Opus-4.6 & 0.775 [0.69,0.85] & 0.589 & 0.812 \\
\sapi~Claude-Opus-4.7 & 0.812 [0.72,0.89] & 0.329 & 0.688 \\
\sapi~Claude-Opus-4.8 & 0.787 [0.70,0.88] & 0.389 & 0.713 \\
\sapi~Gemini-2.5-Flash & 0.835 [0.73,0.91] & 0.344 & 0.759 \\
\sapi~Gemini-2.5-F.Lite & 0.818 [0.71,0.90] & 0.324 & 0.753 \\
\sapi~Gemini-2.5-Pro & 0.838 [0.76,0.91] & 0.372 & 0.762 \\
\sapi~Gemini-3-Flash-Pv & 0.787 [0.69,0.88] & 0.482 & 0.750 \\
\sapi~Gemini-3.1-Lite & 0.812 [0.74,0.90] & 0.453 & 0.750 \\
\sapi~Gemini-3.1-Pro-Pv & 0.821 [0.73,0.88] & 0.478 & 0.833 \\
\sapi~Gemini-3.5-Flash & 0.823 [0.73,0.90] & 0.436 & 0.772 \\
\bottomrule
\end{tabular}

\end{table}

\begin{table}[htb]
  \centering\scriptsize\setlength{\tabcolsep}{3pt}
  \caption{Foundation-model parse failures (null responses) per task; encoders always
  produce a response.
  Nulls are excluded from scores, never counted as errors.}
  \label{tab:supp-parse}
  \begin{tabular}{@{}lrrrrr@{}}
\toprule
model & silh. & trip. & series & ooo & total \\
\midrule
SmolVLM-256M & 1 & 0 & 0 & 0 & 1/740 (0.1\%) \\
SmolVLM-2B & 1 & 0 & 0 & 0 & 1/740 (0.1\%) \\
Qwen2-VL-2B & 6 & 0 & 0 & 0 & 6/740 (0.8\%) \\
LLaVA-1.5-7B & 0 & 0 & 0 & 0 & 0/740 (0.0\%) \\
LLaVA-1.6-7B & 3 & 0 & 0 & 2 & 5/740 (0.7\%) \\
Qwen2-VL-7B & 1 & 0 & 0 & 4 & 5/740 (0.7\%) \\
Idefics2-8B & 0 & 0 & 0 & 0 & 0/740 (0.0\%) \\
GPT-5 & 1 & 0 & 2 & 7 & 10/740 (1.4\%) \\
GPT-5-Mini & 0 & 0 & 0 & 0 & 0/740 (0.0\%) \\
GPT-5-Nano & 1 & 8 & 2 & 7 & 18/740 (2.4\%) \\
GPT-5.1 & 0 & 0 & 0 & 0 & 0/740 (0.0\%) \\
GPT-5.2 & 0 & 0 & 0 & 0 & 0/740 (0.0\%) \\
GPT-5.4 & 0 & 0 & 0 & 0 & 0/740 (0.0\%) \\
GPT-5.4-Pro & 6 & 0 & 0 & 2 & 8/740 (1.1\%) \\
GPT-5.5 & 0 & 3 & 1 & 0 & 4/740 (0.5\%) \\
Claude-Haiku-4.5 & 0 & 0 & 0 & 0 & 0/740 (0.0\%) \\
Claude-Sonnet-4.5 & 0 & 0 & 0 & 0 & 0/740 (0.0\%) \\
Claude-Sonnet-4.6 & 0 & 0 & 0 & 0 & 0/740 (0.0\%) \\
Claude-Sonnet-5 & 0 & 0 & 0 & 0 & 0/740 (0.0\%) \\
Claude-Opus-4.5 & 0 & 0 & 0 & 0 & 0/740 (0.0\%) \\
Claude-Opus-4.6 & 0 & 0 & 0 & 0 & 0/740 (0.0\%) \\
Claude-Opus-4.7 & 0 & 0 & 0 & 0 & 0/740 (0.0\%) \\
Claude-Opus-4.8 & 0 & 0 & 0 & 0 & 0/740 (0.0\%) \\
Gemini-2.5-Flash & 1 & 0 & 2 & 3 & 6/740 (0.8\%) \\
Gemini-2.5-F.Lite & 3 & 0 & 0 & 0 & 3/740 (0.4\%) \\
Gemini-2.5-Pro & 0 & 0 & 0 & 0 & 0/740 (0.0\%) \\
Gemini-3-Flash-Pv & 0 & 0 & 1 & 0 & 1/740 (0.1\%) \\
Gemini-3.1-Lite & 0 & 0 & 0 & 0 & 0/740 (0.0\%) \\
Gemini-3.1-Pro-Pv & 2 & 0 & 0 & 0 & 2/740 (0.3\%) \\
Gemini-3.5-Flash & 1 & 1 & 0 & 1 & 3/740 (0.4\%) \\
\bottomrule
\end{tabular}

\end{table}

\end{document}